\documentclass{article}

\usepackage[final, nonatbib]{neurips_wrl2022}

\usepackage[utf8]{inputenc} %
\usepackage[T1]{fontenc}    %
\usepackage{hyperref}       %
\usepackage{url}            %
\usepackage{booktabs}       %
\usepackage{amsfonts}       %
\usepackage{nicefrac}       %
\usepackage{microtype}      %

\usepackage{multirow}
\usepackage{subcaption}
\usepackage{graphicx}
\usepackage{xr}
\title{A Benchmark for Out of Distribution Detection in Point Cloud 3D Semantic Segmentation}

\author{%
    \parbox{\linewidth}{
        Lokesh Veeramacheneni$^1$, Matias Valdenegro-Toro$^2$\\
    }\\
    ~\\
    \parbox{\linewidth}{
        $^1$ Bonn-Rhein-Sieg University of Applied Sciences, 53757 Sankt Augustin, Germany.\\
        $^2$ Department of AI, University of Groningen, 9747 AG Groningen, The Netherlands.\\
        ~\\
        \texttt{lokesh.Veeramacheneni@smail.inf.h-brs.de}, \, \texttt{m.a.valdenegro.toro@rug.nl}
    }
}

\begin{document}

\maketitle

\begin{abstract}

  Safety-critical applications like autonomous driving use Deep Neural Networks (DNNs) for object detection and segmentation.
  The DNNs fail to predict when they observe an Out-of-Distribution (OOD) input leading to catastrophic consequences.
  Existing OOD detection methods were extensively studied for image inputs but have not been explored much for LiDAR inputs.
  So in this study, we proposed two datasets for benchmarking OOD detection in 3D semantic segmentation.
  We used Maximum Softmax Probability and Entropy scores generated using Deep Ensembles and Flipout versions of RandLA-Net as OOD scores.
  We observed that Deep Ensembles out perform Flipout model in OOD detection with greater AUROC scores for both datasets.
\end{abstract}
\section{Introduction}
Deep Neural Networks (DNNs) have made tasks such as object detection and classification easy to solve, allowing deployment in various real-world scenarios such as autonomous driving, and robotic surgery. These networks are usually trained and evaluated on similar datasets.
This is called the closed world assumption, where the train and test sets have the same classes and similar distributions for inputs and labels. But this assumption often does not hold in practice, requiring out of distribution detection, in order for safe use of perception models in the real world. If a model is uncertain, it can abstain to provide an answer and could trigger additional processing or release control to a human.

Semantic segmentation is a popular way to understand a scene. Since autonomous vehicles and other kinds of robots operate in 3D environment, point clouds are preferred to perform perception, leading into 3D semantic segmentation, where each point in the cloud receives a class label.

This paper introduces the problem of out of distribution detection in 3D semantic segmentation tasks. We propose a benchmark using two well known LiDAR datasets \cite{survey3d} (Semantic3D and S3DIS) in order to benchmark out of distribution detection capabilities. We initially benchmark uncertainty-based methods, namely Dropout, Deep Ensembles, and Flipout, using RandLA-Net \cite{Hu_2020_CVPR_Randla} as model for 3D semantic segmentation. Our benchmark contains two dataset combinations, namely Semantic3D vs S3DIS (Benchmark A), and Semantic3D vs Semantic3D without color information (Benchmark B).

Our initial results indicate that OOD detection in point cloud segmentation is possible, with the best result given by Deep Ensembles (AUROC 0.893) with maximum probability on benchmark A, and 0.773 on Benchmark B.

The contributions of this paper are: a definition of a benchmark for out of distribution detection in semantic segmentation of point clouds, initial benchmark results for uncertainty-based methods (Dropout, Ensembles, Flipout), using both entropy and maximum probability.

\section{Related Work}

\subsection{3D Semantic Segmentation}
\label{subsec:semseg}
Existing 3D semantic segmentation can be broadly grouped into three different kinds.
The first type includes point-based models where the model directly feeds on a 3D point cloud.
Pointnet \cite{Qi_2017_CVPR_pointnet}, Pointnet++ \cite{qi2017pointnet++}, SPLATNet \cite{Su_2018_CVPR_splatnet} and RandLA-Net \cite{Hu_2020_CVPR_Randla} are few example networks of this type.
The second type includes projection-based models where the data is projected onto a 2D range image.
SqueezeSegV3 \cite{xu2020squeezesegv3}, RangeNet++ \cite{Milioto2019}, KPRNet \cite{kochanov2020kprnet} and 3DMiniNet \cite{3Dmininet} are few example models for this type.
Projection-based models also include projecting onto bird eye view projection of point cloud.
SalsaNext \cite{SalsaNext_2020}, PolarNet \cite{polarnet} and Cylinder3D \cite{zhu2020cylindrical} are few models are of this subtype.

The final type of model is graph neural networks.
Dynamic graph CNN \cite{dyn_graph_cnn} and GACNet \cite{Wang_2019_CVPR_GACNet} are example models using graph neural networks.
In this study, we use RandLA-Net \cite{Hu_2020_CVPR_Randla} which is point based model with fewer parameters and also state-of-the-art performance in point-based methods without expensive operations like kernelization.
Moreover, RandLA-Net doesn't require preprocessing like range image computation or farthest point sampling and employs simple random point sampling. 

\subsection{Out-of-Distribution detection}
\label{subsec:ood}
A dataset is referred to as the OOD dataset if the whole dataset consists of only OOD objects and the training dataset is called In-Distribution (ID) dataset.
Multiple approaches exist to generate the scores for OOD detection.
These approaches employ a threshold-based method and have no idea of how OOD data is during training.
\cite{hendrycks2016baseline_MSP} provides a baseline method using Maximum Softmax Probability (MSP) scores for OOD detection.
An improved method called ODIN for the baseline method is proposed in \cite{liang2017enhancing_ODIN}.
ODIN utilizes calibrated softmax scores along with input noise perturbations making training adversarial.
\cite{lee2018simple_mahalanobis} proposed the use of Mahalanobis distance as OOD score instead of softmax, where Mahalanobis distance is calculated between each activation map and multivariate Gaussian distribution.
\cite{Maha_plus_ODIN} uses the combination of Mahalanobis distance and ODIN for OOD score generation.
\cite{ReAct} proposed a method for OOD score generation called ReAct, where the proposed ReAct activation is applied before the softmax layer to suppress the higher activations to a constant.

Based on the fact that the OOD data has higher uncertainty scores when compared to ID data, \cite{lakshminarayanan2016simple} estimates epistemic uncertainty using Deep Ensembles to classify ID and OOD. 
Similarly \cite{JAmersfoot_RBF} uses epistemic uncertainty calculated from radial basis function to detect OOD data.
\cite{UOOD_BNN}, \cite{Grad_UOOD} utilizes Bayesian neural networks for uncertainty estimation and then for OOD detection.

In this paper, we use the MSP score proposed in \cite{hendrycks2016baseline_MSP} with uncertainty estimates from Deep Ensembles \cite{lakshminarayanan2016simple} and Flipout \cite{Flipout} for OOD score generation. Additionally we use Entropy to integrate information from the whole predictive distribution. Both MSP and Entropy metrics are computed from the output predictive distribution, which is the average of ensemble/forward pass outputs of each network. The selection of UQ methods is motivated as these methods are scalable \cite{gustafsson2020evaluating} and have been tested to work well for point cloud semantic segmentation \cite{bhandary2020evaluating}.

\section{Datasets and Benchmark}
\label{sec:dataset}
All the experiments discussed in this paper use Semantic3D proposed in \cite{hackel2017semantic3d} as the In-Distribution dataset.
Semantic3D is chosen as the ID dataset because it is one of the dense datasets along with the RGB color.
We hypothesize that having RGB colors will help in the improved performance of OOD detection.

The first OOD dataset we used is S3DIS proposed in \cite{Armeni_2016_CVPR_S3DIS}.
We especially chose the S3DIS as the OOD dataset because the dataset consists of indoor objects whereas the ID dataset consists of outdoor objects. This domain difference in scenes makes S3DIS an ideal OOD dataset.
We expect the Semantic3D trained RandLA-Net model to detect the S3DIS dataset as OOD with ease and high confidence.
This expectation is due to the difference in point geometry between these datasets. We call this combination Benchmark A.

The second OOD dataset we used is Semantic3D without color. Having this dataset as an OOD dataset ensures the same point geometries between ID and OOD datasets but the difference in other point properties like color. This simulates sensor failure.
Because of the above reason, we expect RandLA-Net to struggle in detecting this OOD data. We call this combination Benchmark B.
\section{Experiments and Results}
\label{sec:experiments}
The experimental setup of RandLA-Net on Deep Ensembles and Flipout is reported in Supplementary Section~\ref{sec:exp_setup}.
Training results (mean IoU, per-class Iou, Overall Accuracy) are also provided in Supplementary Section~\ref{sec:randla_eval}.
Here we discuss the results of OOD detection on two proposed datasets using the AUROC score as the evaluation metric.
\subsection*{Benchmark A - Semantic3D vs S3DIS }
In this section, we evaluate the OOD detection performance on the Semantic3D-vs-S3DIS dataset with Semantic3D being ID and S3DIS being the OOD dataset.
We use the AUROC scores generated using Maximum Softmax Probability and Entropy from Deep Ensembles, Flipout and Dropout techniques.
From AUROC (Dataset1) column in Table~\ref{tab:sem3dvs3dis_auroc}, we observe that the Deep Ensembles outperform Flipout and Dropout models in OOD detection, with Flipout only outperforming the other two methods in the case of a single ensemble/forward pass, which indicates an advantage of the Gaussian weight distribution learned using Flipout.
We also observe that after the Ensemble size or number of passes 10, the performance improvements are little to none.

\begin{figure}
    \begin{subfigure}{0.45\textwidth}
        \centering
        \includegraphics[scale=0.39]{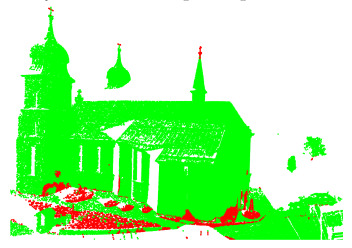}
        \caption{Semantic3D - ID}
        \label{fig:bin_id_ood_sem3d_DE}
    \end{subfigure}
    \begin{subfigure}{0.45\textwidth}
        \centering
        \includegraphics[scale=0.39]{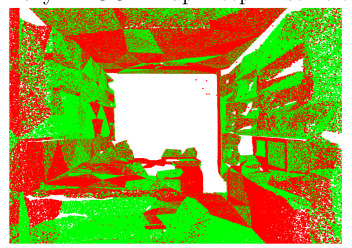}
        \caption{S3DIS - OOD}
        \label{fig:bin_id_ood_s3dis_DE}
    \end{subfigure}
    \caption{Images depicting the ID points in green and OOD points in red for Semantic3D (ID) dataset in (a) and S3DIS (OOD) dataset in (b).
    ID-OOD classification is made using Maximum Softmax Probability values generated from Deep Ensembles with ensemble size of 10.}
\end{figure}

Figure~\ref{fig:bin_id_ood_sem3d_DE} and \ref{fig:bin_id_ood_s3dis_DE} depict the ID points represented in green and OOD points represented in red for both Semantic3D (ID) and S3DIS (OOD) datasets respectively.
Here we observe that a few points such as the edges of the church in Figure~\ref{fig:bin_id_ood_sem3d_DE} are classified with a low probability score and these points are also classified as OOD points in the ID dataset.
Similarly, a few points in S3DIS dataset in Figure~\ref{fig:bin_id_ood_s3dis_DE} are also classified as a part of ID dataset (green points).
These green points are because of the walls being classified as a building which is partly true.
Overall, we observe that the S3DIS dataset has higher OOD points (more reddish) whereas most of the ID dataset is greenish in color.
 
\begin{table}
    \centering
    \begin{tabular}{llllll}
    \toprule
    Ensemble size/ \#passes & Method               &  \multicolumn{2}{c}{AUROC Benchmark A} &  \multicolumn{2}{c}{AUROC Benchmark B}          \\
    \midrule
                            &                      &  MSP              & Entropy           &  MSP              & Entropy\\ \hline
    \multirow{3}{*}{1}      & Dropout              & 0.53311          & 0.53041            & 0.66349          & 0.65908\\
                            & Flipout              & \textbf{0.69988} & \textbf{0.69368}   & 0.64221          & 0.66157\\
                            & Deep Ensembles       & 0.62020          & 0.62529            & \textbf{0.67855} & \textbf{0.67866}\\
    \midrule
    \multirow{3}{*}{5}      & Dropout              & 0.58439          & 0.57821            & 0.69448          & 0.68507\\
                            & Flipout              & 0.77885          & 0.76934            & 0.63743          & 0.66536\\
                            & Deep Ensembles       & \textbf{0.84013} & \textbf{0.83665}   & \textbf{0.76769} & \textbf{0.77120}\\
    \midrule
    \multirow{3}{*}{10}     & Dropout              & 0.60168          & 0.59925            & 0.68568          & 0.68004\\
                            & Flipout              & 0.78728          & 0.78327            & 0.63712          & 0.66535 \\
                            & Deep Ensembles       & \textbf{0.87929} & \textbf{0.87541}   & \textbf{0.77837} & \textbf{0.78142}\\
    \midrule
    \multirow{3}{*}{15}     & Dropout              & 0.59773          & 0.59557            & 0.68975          & 0.68347\\
                            & Flipout              & 0.7667           & 0.76741            & 0.63022          & 0.65976\\
                            & Deep Ensembles       & \textbf{0.88486} & \textbf{0.88246}   & \textbf{0.77302} & \textbf{0.77881}\\
    \midrule
    \multirow{3}{*}{20}     & Dropout              & 0.59766          & 0.59661            & 0.68447          & 0.68199 \\
                            & Flipout              & 0.77331          & 0.77237            & 0.63017          & 0.65934\\
                            & Deep Ensembles       & \textbf{0.89338} & \textbf{0.89052}   & \textbf{0.77031} & \textbf{0.77584}\\
    \bottomrule
    \end{tabular}
    \caption{AUROC scores calculated for all the points in the test sets of both datasets. MSP and Entropy values are represented for Deep Ensembles, Flipout and Dropout methods with increase in ensemble size or number of passes with step size of 5.
    Benchmark A represents Semantic3D vs S3DIS and Benchmark B is Semantic3D vs Semantic3D without color.}
    \label{tab:sem3dvs3dis_auroc}
\end{table}

\subsection*{Benchmark B - Semantic3D vs Semantic3D without color}
\begin{figure}
    \begin{subfigure}{0.45\textwidth}
        \centering
        \includegraphics[scale=0.40]{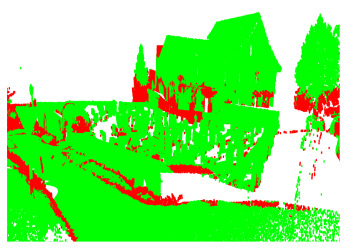}
        \caption{Semantic3D - ID}
        \label{fig:bin_id_ood2_sem3d_DE}
    \end{subfigure}
    \begin{subfigure}{0.45\textwidth}
        \centering
        \includegraphics[scale=0.40]{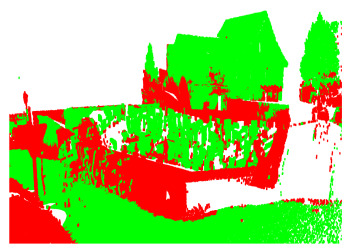}
        \caption{Semantic3D without color - OOD}
        \label{fig:bin_id_ood2_sem3dwoc_DE}
    \end{subfigure}
    \caption{Images depicting the ID points in green and OOD points in red for Semantic3D (ID) dataset in (a) and Semantic3D without color (OOD) dataset in (b).
    ID-OOD classification is made using Maximum Softmax Probability values generated from Deep Ensembles with ensemble size of 10.}
\end{figure}

Similarly, the second set of AUROC columns in Table~\ref{tab:sem3dvs3dis_auroc} represents the AUROC scores generated using MSP and Entropy for the second OOD benchmark.
Deep Ensembles outperform Flipout and Dropout in this dataset also.
In both the dataset performances, we observe that the AUROC scores for the Entropy and MSP are similar.
Performance of OOD detection maxes out at ensemble size or the number of passes of 10 and no improvements are observed further increase in size.
Figures~\ref{fig:bin_id_ood2_sem3d_DE} and \ref{fig:bin_id_ood2_sem3dwoc_DE} depict the ID and OOD points (green and red color respectively) for Semantic3D and Semantic3D without color respectively.
Most of the misclassified points in the ID dataset are classified as OOD points as shown in Figure~\ref{fig:bin_id_ood2_sem3d_DE}.
In Semantic3D without color (OOD dataset) we majorly observe that walls and low vegetation are prone to misclassify as buildings and manmade terrain because no color information is available to differentiate between them.
So most of the points belonging to walls and low vegetation are detected as OOD points.

\section{Conclusions}
In this paper, we studied the performance of OOD detection in 3D semantic segmentation, by proposing two benchmarks for OOD benchmarking, one being Semantic3D-vs-S3DIS and the other dataset being Semantic3D-vs-Semantic3D without color.
We trained a RandLA-Net model for 3D semantic segmentation on the Semantic3D dataset and ran inference on S3DIS and Semantic3D without color datasets individually.
Overall, we observe that the OOD detection performance using Deep Ensembles is better than two other Bayesian methods.
Finally, we conclude that OOD detection is relatively easy in case the OOD objects varying point geometry compared to training data.
In case the OOD object has similar point geometry to training data, it is challenging as the RandLA-Net model hugely relies on point geometries. Additionally our results show that OOD detection is still challenging even in a simple setup like Benchmark A, requiring large computational costs. Future work should produce lightweight uncertainty quantification methods.

\bibliographystyle{plain}
\bibliography{bibliography.bib}

\newpage
\appendix
\section*{\centering\Large{Appendix}}

\section{Experimental Setup}
\label{sec:exp_setup}
\subsection*{Deep Ensembles}
We trained 20 instances of RandLA-Net by the same procedure as the authors described in \cite{Hu_2020_CVPR_Randla}.
The only change is made during the inference, we changed the inference pipeline to infer over all the points in the test set.
The outputs of these 20 instances are averaged to extract the entropy and MSP values.
\subsection*{Flipout}
We changed the last three classification layers of the RandLA-Net model to Flipout compatible using Tensorflow-probability API \cite{tf_Probability}.
After exhaustive trial and error, the prior for Flipout layers is chosen as normal prior with unit standard deviation and zero mean.
The same training hyperparameters as Deep Ensembles are used to train the Flipout style RandLA-Net.
We used 20 inferences from this network to extract the entropy and MSP values.
\subsection*{Dropout}
The RandLA-Net architecture proposed in \cite{Hu_2020_CVPR_Randla} has a Dropout layer at the end of the classification layers.
We also used the same setup with 0.5 as Dropout probability.
Similar to the Flipout model we used 20 inferences to extract entropy and MSP values.

\section{OOD Detection Evaluation - AUROC Computation}
In this section we describe how AUROC is computed for out of distribution detection in point cloud segmentation predictions.

We first computed the softmax predictions on each point in the test set of the ID dataset and similarly softmax predictions on each point in the test set of the OOD dataset. Then we computed the MSP and Entropy values from each point predictions in both ID and OOD datasets. These scores are used to compute AUROC metrics.

The computed per point MSP and Entropy values are used to compute the AUROC using the sklearn API. Each point in a ID point cloud receives a label $y = 0$, while each point in an OOD point cloud receives a label $y = 1$. AUROC is computed between all the point clouds in a given dataset combination. This is done because the number of points in each cloud is large, but the number of point cloud samples in each dataset is relatively small.

This means that the AUROC results presented in our paper indicate the ability of a model to discriminate individual points in the cloud as in-distribution or out of distribution. This is slightly different than the standard case of classification, where a whole sample is classified in the same way.

\section{RandLA-Net Evaluation}
\label{sec:randla_eval}
\subsection*{Deep Ensembles}
\begin{figure}[h!]
    \centering
    \includegraphics[scale=0.45]{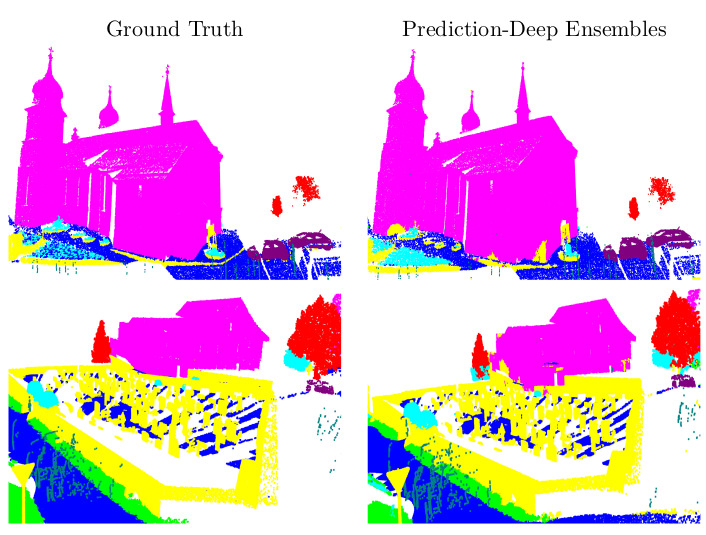}
    \includegraphics[scale=0.33]{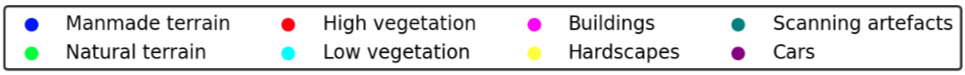}
    \caption{Predictions of RandLA-Net Deep Ensembles on Semantic3d dataset with ground truth in left column and predictions on right column.
    Predictions are computed with an ensemble size of 15.}
    \label{fig:semantic3d_DE_output}
\end{figure}

In this section, we discuss the training results of RandLA-Net over 20 Deep Ensembles.
Table~\ref{tab:ensemble_eval} enumerates the meanIoU, per-class IoU, and Accuracy of RandLA-Net up to an ensemble size of 20.
From the table, we observe that there is an improvement in performance with the increase in ensemble size.
Also after the ensemble size of 10, the gains are very little to none.
Figure~\ref{fig:semantic3d_DE_output} depicts the predictions of RandLA-Net with an ensemble size of 15.
Typically, we observed that there is some amount of misclassification along the edges of the church, and trees.
\begin{table}[h!]
    \resizebox{\textwidth}{!}{%
    \begin{tabular}{lllllllllll}
    \toprule
    & & \multicolumn{7}{c}{\textbf{IoU per-class}} & \\
    \midrule
    \textbf{Ensemble size} & \textbf{meanIoU} & \textbf{C1} & \textbf{C2} & \textbf{C3} & \textbf{C4} & \textbf{C5} & \textbf{C6} & \textbf{C7} & \textbf{C8} & \textbf{Accuracy} \\
    \midrule
    1& 68.19& 94.55& 81.19& 84.67& 29.43& 81.37& 18.85& 64.74& 90.74& 88.78 \\
    5& 69.51& 94.73& 81.92& 84.42& 28.05& \textbf{86.41}& 28.50& 61.03& 91.03& 90.04 \\
    10& 69.97& 95.25& 83.73& 86.63& 30.36& 84.13& 18.60& \textbf{66.01}& 92.61& 89.94 \\
    15& 70.32& 95.27& 83.54& \textbf{88.22}& \textbf{32.19}& 84.82& 26.17& 61.67& 90.75& \textbf{90.57} \\
    20& \textbf{70.80}& \textbf{95.55}& \textbf{84.11}& 86.65& 29.60& 85.41& \textbf{29.58}& 62.47& \textbf{93.06}& 90.56 \\
    \bottomrule
    \end{tabular}%
    }
    \caption{Illustration of performance of RandLA-Net on Semantic3D over ensemble size. meanIOU, IOU per-class and overall accuracy are represented here.
    C1 to C8 are the classes of Semantic3D which are Manmade terrain, Natural terrain, High vegetation, Low vegetation, Buildings, Hardscapes, Scanning artefacts, and Cars.}
    \label{tab:ensemble_eval}
\end{table}

\subsection*{Flipout}
\begin{figure}
    \centering
    \includegraphics[scale=0.4]{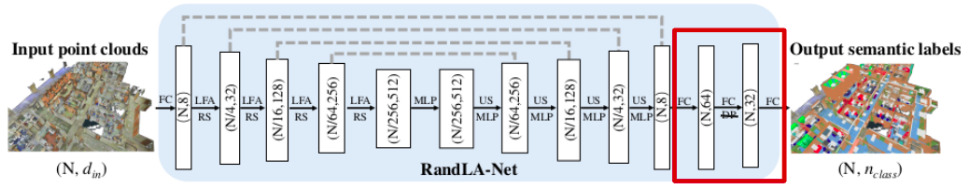}
    \caption{Flipout style RandLA-Net where the last three FC layers as depicted in red box are made
    Flipout-compatible.}
    \label{randlanet_fout_layers}
\end{figure}

The following three layers highlighted in the red box in Figure~\ref{randlanet_fout_layers} are changed to Flipout compatible.
Table~\ref{tab:flipout_eval} depicts the mean IoU, per-class IoU and Accuracy of the Semantic3D trained Flipout style RandLA-Net.
Even though the meanIoU, Accuracy and most of the classes IoU are similar to the results produced in \cite{Hu_2020_CVPR_Randla}
\footnote[1]{Since Semantic3D is an ongoing benchmark challenge, the test set is not public. We separated a part of the training set and used it as a test set. For reference, we evaluated the RandLA-Net on this test set with the weights provided by the authors.}.
Here we observe a significant improvement in the Hardscapes class (C6 in table) compared to Deep Ensembles.
Unlike Deep Ensembles, with the increase in the number of forward passes we do not observe any performance improvements and judging by meanIoU nd Accuracy, one can say that there is a slight decrease in overall performance with increase in the number of passes.
\begin{figure}[h!]
    \centering
    \includegraphics[scale=0.45]{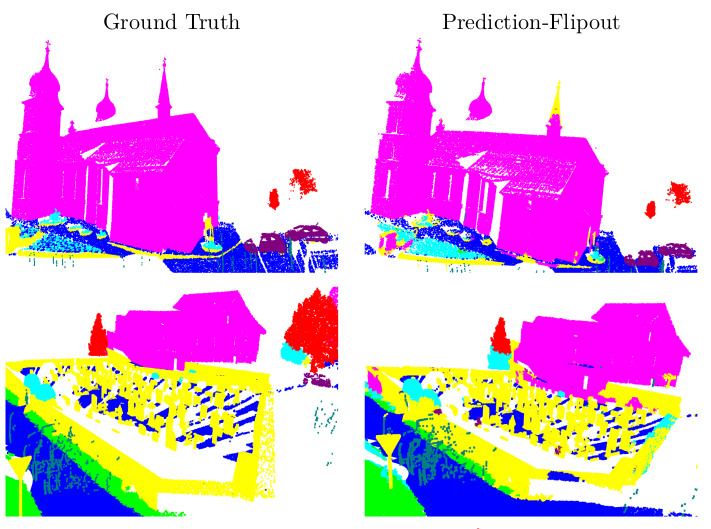}
    \includegraphics[scale=0.33]{images/legend.jpg}
    \caption{Predictions of Flipout style RandLA-Net on Semantic3d dataset with ground truth in left column and predictions on right column.
    Predictions are computed with 15 passes.}
    \label{fig:semantic3d_Fout_output}
\end{figure}
\begin{table}[h!]
    \resizebox{\textwidth}{!}{%
    \begin{tabular}{lllllllllll}
    \toprule
    & & \multicolumn{7}{c}{\textbf{IoU per-class}} & \\ 
    \midrule
    \textbf{\#Passes} & \textbf{MeanIoU} & \textbf{C1} & \textbf{C2} & \textbf{C3} & \textbf{C4} & \textbf{C5} & \textbf{C6} & \textbf{C7} & \textbf{C8} & \textbf{Accuracy} \\
    
    1& 69.95  & 94.24&80.09&86.16&22.48&88.70&39.41&57.42&91.12&90.71\\
    5& 69.83  & 94.38&80.21&84.10&23.32&87.80&39.68&57.75&91.43&90.43\\
    10& 69.84 & 94.38&80.16&83.90&23.46&87.73&39.75&57.83&91.47&90.40\\
    15& 69.86 & 94.38&80.17&83.80&23.48&87.73&39.82&57.96&91.57&90.40\\
    20& 69.87 & 94.38&80.18&83.80&23.57&87.72&39.84&57.92&91.57&90.40\\
    \bottomrule
    \end{tabular}%
    }
    \caption{Illustration of performance of Flipout style RandLA-Net on Semantic3D dataset. meanIOU, IOU per-class and overall accuracy are represented here.
    C1 to C8 are the classes of Semantic3D which are Manmade terrain, Natural terrain, High vegetation, Low vegetation, Buildings, Hardscapes, Scanning artefacts, and Cars.}
    \label{tab:flipout_eval}
\end{table}

\section{OOD Dataset Predictions}
\begin{figure}[h!]
    \centering
    \includegraphics[scale=0.4]{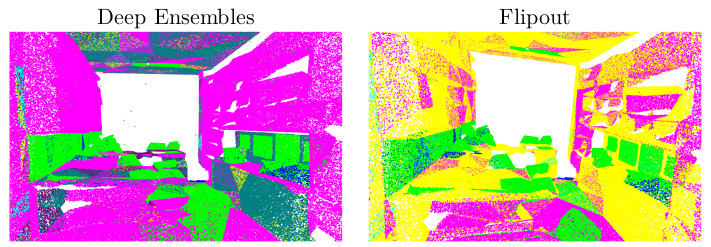}
    \includegraphics[scale=0.41]{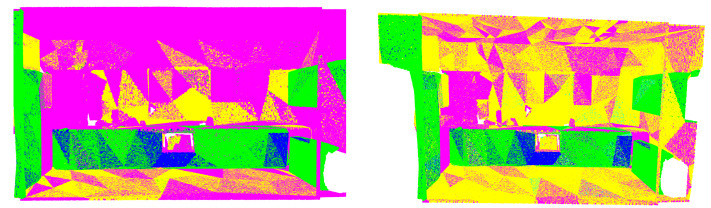}
    \includegraphics[scale=0.3]{images/legend.jpg}
    \caption{Predictions of RandLA-Net over S3DIS (OOD) dataset with first column using Deep Ensembles and second column using Flipout.
    Predictions are computed using ensemble size and number of passes as 15.}
    \label{fig:s3dis_predictions}
\end{figure}

Figure~\ref{fig:s3dis_predictions} shows the predictions of the RandLA-Net model on the S3DIS dataset (first OOD dataset).
The first column represents the predictions on Deep Ensembles size of 15 and the second column with predictions from the Flipout model with 15 forward passes.
In the case of Deep Ensembles, we observe that most of the walls are segmented as the building which is partly true but the other objects like cabinets, printer/wall posters, and chairs are misclassified.
Whereas in the case of the Flipout model, most of the points are predicted as Hardscapes but both the Deep Ensembles and Flipout classify the cabinets and printer/wall posters as low vegetation.
This is mostly because the feature vectors are near the feature vector of low vegetation.
A prominent observation is that the predictions on the OOD dataset are triangular this is because of the inherent property of the data.
S3DIS dataset is generated using a matterport scanner and according to \cite{matterport_scan_generation} the matterport camera first generates a triangular mesh and then the points are extracted from these meshes.

\section{ID-OOD Maps - Flipout}
\begin{figure}
    \begin{subfigure}{0.45\textwidth}
        \centering
        \includegraphics[scale=0.5]{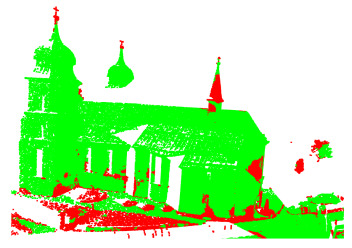}
        \caption{Semantic3D - ID}
        \label{fig:bin_id_ood_sem3d_fout}
    \end{subfigure}
    \begin{subfigure}{0.45\textwidth}
        \centering
        \includegraphics[scale=0.5]{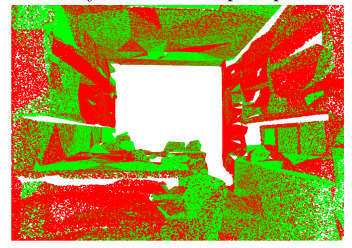}
        \caption{S3DIS - OOD}
        \label{fig:bin_id_ood_s3dis_fout}
    \end{subfigure}
    \caption{Images depicting the ID points in green and OOD points in red for Semantic3D (ID) dataset in (a) and S3DIS (OOD) dataset in (b).
    ID-OOD classification is made using Maximum Softmax Probability values generated from Flipout with 10 passes.}
\end{figure}

\begin{figure}
    \begin{subfigure}{0.45\textwidth}
        \centering
        \includegraphics[scale=0.5]{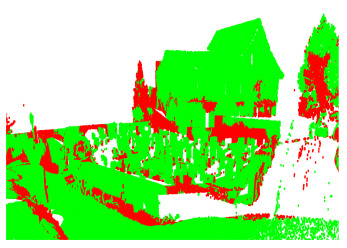}
        \caption{Semantic3D - ID}
        \label{fig:bin_id_ood2_sem3d_fout}
    \end{subfigure}
    \begin{subfigure}{0.45\textwidth}
        \centering
        \includegraphics[scale=0.5]{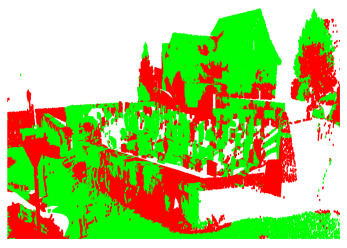}
        \caption{Semantic3D without color - OOD}
        \label{fig:bin_id_ood2_sem3dwoc_fout}
    \end{subfigure}
    \caption{Images depicting the ID points in green and OOD points in red for Semantic3D (ID) dataset
    in (a) and Semantic3D without color (OOD) dataset in (b). ID-OOD classification is made using Maximum Softmax
    Probability values generated from Flipout with 10 passes}
\end{figure}

In this section, we present the ID and OOD points classified using threshold from MSP values extracted using Flipout style RandLA-Net with 10 passes.
Figures~\ref{fig:bin_id_ood_sem3d_fout} and \ref{fig:bin_id_ood_s3dis_fout} represent the ID points in green and OOD points in red for the first dataset (Semantic3D-vs-S3DIS).
Similarly Figures~\ref{fig:bin_id_ood2_sem3d_fout} and \ref{fig:bin_id_ood2_sem3dwoc_fout} represent ID and OOD points for the second dataset.

\section{OOD Detection ROC Curves}
In this section, we depict the ROC curves along with the optimal threshold MSP and Entropy values extracted from ROC curves.
For both datasets, we use ensemble size and the number of passes of 10.
\subsection*{Semantic3D-vs-S3DIS}
\begin{figure}
        \centering
        \includegraphics[scale=0.4]{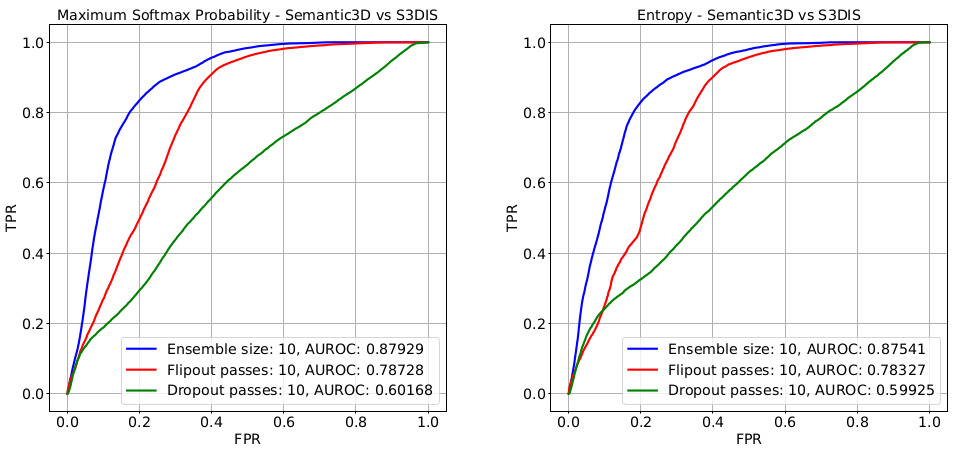}
        \caption{ROC curves of Semantic3D-vs-S3DIS for 10 Ensembles, 10 forward passes for Flipout and
        Dropout. Left image represents the ROC curves using Maximum Softmax Probability and right image represents ROC curves using Entropy.}
        \label{fig:roc_ood1}
    \end{figure}

Figure~\ref{fig:roc_ood1} depicts the ROC curves for both MSP and Entropy for the first OOD dataset respectively.
We observe that the Deep Ensembles' ROC curve is higher than other methods.
Table~\ref{tab:thresholds} represents the extracted optimal thresholds from the above ROC curves.
    \begin{table}[h!]
        \centering
        \begin{tabular}{llll}
            \toprule
            OOD Benchmark                                           & Method    & MSP   & Entropy \\
            \midrule
            \multirow{2}{*}{Semantic3D-vs-S3DIS}                    & Ensembles & 0.755 & 0.386   \\
                                                                    & Flipout   & 0.717 & 0.439   \\
            \bottomrule
        \end{tabular}
        \caption{MSP threshold and entropy threshold generated from the corresponding ROC curves for Semantic3D-vs-S3DIS datasets for MSP and entropy methods.}
        \label{tab:thresholds}
    \end{table}
\subsection*{Semantic3D-vs-Semantic3D without color}
\begin{figure}[t]
    \centering
    \includegraphics[scale=0.4]{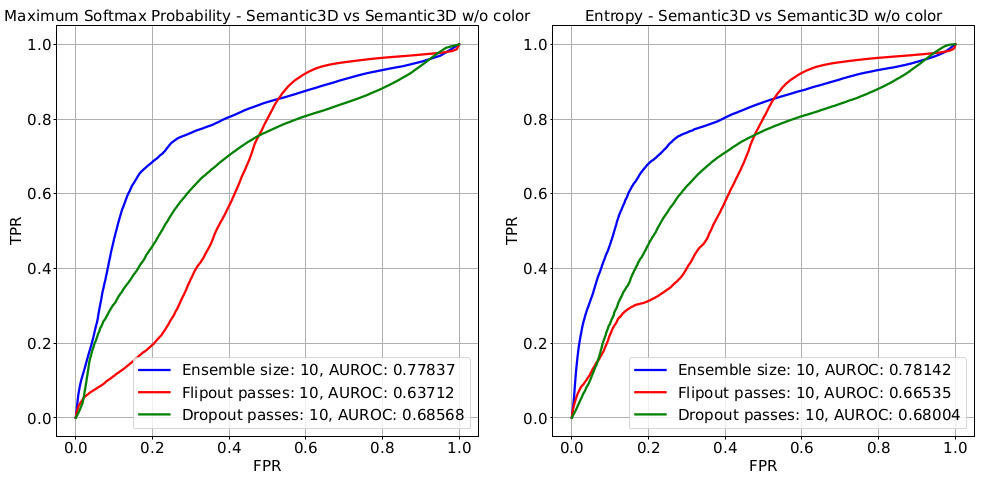}
    \caption{ROC curves of Semantic3D-vs-Semantic3D without color for 10 Ensembles, 10 forward passes for Flipout and
    Dropout. Left image represents the ROC curves using Maximum Softmax Probability and right image represents ROC curves using Entropy.}
    \label{fig:roc_ood2}
\end{figure}

Similarly, Figure~\ref{fig:roc_ood2} represents the ROC curves for the second OOD dataset along with thresholds represented in Table~\ref{tab:thresholds_2}.
Here we observe that the ROC curves are smaller than the first OOD dataset.
Also, the thresholds for MSP are higher and for Entropy are much lower.
This kind of behaviour is expected because Semantic3D without color dataset has the same point geometry as the ID dataset.
Whereas in the previous case, we observe lower thresholds for MSP and higher thresholds for Entropy.
\begin{table}[h!]
    \centering
    \begin{tabular}{llll}
        \toprule
        OOD Benchmark                                           & Method    & MSP   & Entropy \\
        \midrule
       \multirow{2}{*}{Semantic3D-vs-Semantic3D without color} & Ensembles & 0.790 & 0.323   \\ 
                                                               & Flipout   & 0.796 & 0.131   \\
       \bottomrule
    \end{tabular}
    \caption{MSP threshold and entropy threshold generated from the corresponding ROC curves for Semantic3D-vs-Semantic3D without color datasets for MSP and entropy methods.}
    \label{tab:thresholds_2}
\end{table}

\end{document}